\documentclass[runningheads]{llncs}
\usepackage{graphicx}
\begin{document}
\title{BRAIN2DEPTH: Lightweight CNN Model for Classification of Cognitive States from EEG Recordings}
\titlerunning{Brain2Depth:Lightweight CNN Model}
%
\author{ Pankaj Pandey\inst{1} \and
Krishna Prasad Miyapuram \inst{2}}
\authorrunning{Pandey \& Miyapuram}
%
\institute{Computer Science and Engineering,\and
Centre for Cognitive and Brain Sciences, IIT Gandhinagar \\
\email{pankaj.p@iitgn.ac.in;kprasad@iitgn.ac.in}}
\maketitle              
\begin{abstract}
Several Convolutional Deep Learning models have been proposed to classify the cognitive states utilizing several neuro-imaging domains. These models have achieved significant results, but they are heavily designed with millions of parameters, which increases train and test time, making the model complex and less suitable for real-time analysis. This paper proposes a simple, lightweight CNN model to classify cognitive states from Electroencephalograph (EEG) recordings. We develop a novel pipeline to learn distinct cognitive representation consisting of two stages. The first stage is to generate the 2D spectral images from neural time series signals in a particular frequency band. Images are generated to preserve the relationship between the neighboring electrodes and the spectral property of the cognitive events. The second is to develop a time-efficient, computationally less loaded, and high-performing model. We design a network containing 4 blocks and major components include standard and depth-wise convolution for increasing the performance and followed by separable convolution to decrease the number of parameters which maintains the tradeoff between time and performance. We experiment on open access EEG meditation dataset comprising expert, nonexpert meditative, and control states. We compare performance with six commonly used machine learning classifiers and four state of the art deep learning models. We attain comparable performance utilizing less than 4\% of the parameters of other models. This model can be employed in a real-time computation environment such as neurofeedback.

\end{abstract}
\section{Introduction}

Deep Learning (DL) has sparked a lot of interest in recent years among various research fields. The most developed algorithm, among several deep learning methods, is the Convolutional Neural network (CNN) \cite{dhillon,zhang}. CNNs have made a revolutionary impact on computer vision, speech recognition, and medical imaging to solve challenging problems that were earlier difficult using traditional techniques. One of the complex problems was classification. In the short span of 4 years from 2012 to 2015, the ImageNet image-recognition challenge, which includes 1000 different classes in 1.2 million images,  has consistently shown reduced error rates from 26\% to below 4\%  using CNN as the major component \cite{khan}. Identification of brain activity using CNNs has established remarkable performance in several brain imaging datasets, including functional MRI, EEG, and MEG. For example, Payan and Montana classified Alzheimer’s and healthy brains with 95\% accuracy using 3D convolution layers on the ADNI public dataset containing 2265 MRI scans \cite{payan}. In the recent work, Dhananjay and colleagues implemented three layers of CNN architecture to predict the song from EEG brain responses with 84\% accuracy, despite having several challenges such as having low SNR, complex naturalistic music, and human perceptual differences \cite{sonawane}. However, all these models are building complex and deep networks without considering the limitation of time and resources. Two main concerns observed with deep and wide architectures are having the millions of parameters that lead to high computational cost and memory requirement. Therefore, this opens up the opportunity to navigate the research to develop lightweight domain-specific architectures for resource and time constraint environments. One such scenario is a real-time analysis of EEG signals.

EEG brain recordings have a high temporal resolution as well as a wide variety of challenges such as low signal-to-noise ratio, noise can be of different shape, for example, artifacts from eye movement, head movement, and electrical line noise \cite{jiang}. Hence, to extract significant features, it requires a sophisticated and efficient method that considers the spatial information in depth. In recent times, deep learning methods have been showing a significant improvement over traditional machine learning algorithms.  And, DL models have been expanding in real-time computation also. Real-time analysis requires fast train as well as test time. The major property that should hold in lightweight architecture is to design the blocks, which reduces trainable parameters while maintaining state-of-the-art performance \cite{wu}. A recent paper\cite{sonawane} on the classification of EEG signals has introduced a significant performance by generating the time-frequency images from EEG signals, but the proposed model is made of several components, which loads the model with 5.8 million parameters. In this study, we have proposed a novel pipeline to generate 2D images from EEG signals and develop a model that produces comparable performance with state-of-the-art networks with minimum time for training and testing and suitable for EEG classification tasks.

\begin{figure}
\centering
\includegraphics[width=8cm,height=3cm]{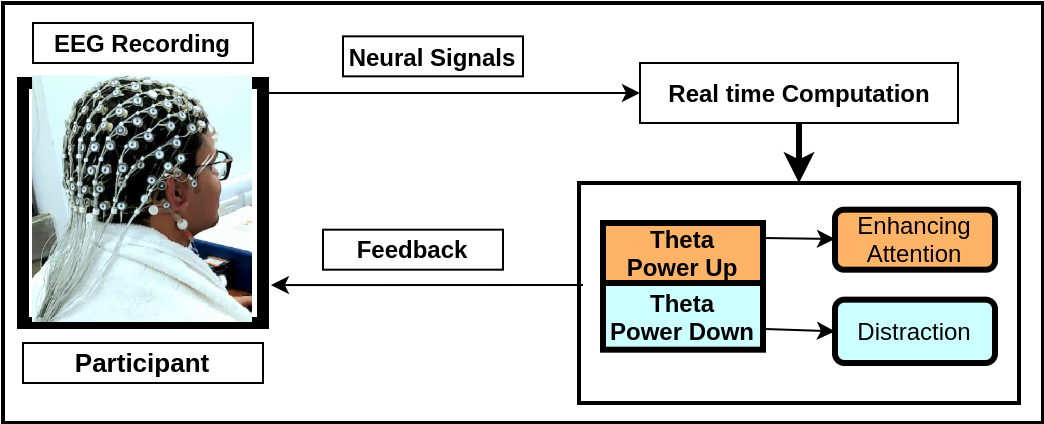}
\caption{NeuroFeedback protocol: EEG signals are passed to the neural computational toolbox, including preprocessing and analysis stages. Analysis pipeline comprises required computational algorithms, which generate feedback for the participant in real-time with minimum lag.}
\label{nf}
\end{figure}

\section{Related Studies}

\subsection{Cognitive Relevance}
Meditation is a mental practice that enhances several cognitive abilities such as enhancing attention, minimizing mind wandering, and developing sustained attention \cite{brandmeyer2}. EEG is the most widely used technique in the neuroscientific study of meditation. EEG signals are decomposed into five frequency bands, including delta (1-4Hz), theta (5-8 Hz), alpha (8-12 Hz), beta (13-30 Hz), and gamma (30-70 Hz). Several meditation studies report the importance of theta and alpha waves for enhancing attention and are associated with several cognitive processes \cite{brandmeyer}. These findings led the researchers to design a neurofeedback protocol \cite{brandmeyer3} to modulate oscillatory activities on the naive participants; a mechanism derives from the focused-attention meditation technique. Neurofeedback is a process to provide visual/audio feedback to a participant while recording his/her neural responses in real-time, which develops skills to self-regulate electrical activity of the brain, as shown in Fig.1. Previous studies on neurofeedback have shown promising results to enhance performance, including athletes (archers improve their shooting performance), musicians, professional dancers, and non-artists to attain skills resembling visual artists \cite{sho}. NeuroFeedback requires a time-efficient and high-performing computational technique in terms to classify the different stages of brain responses.

\subsection{Deep Learning Networks}
There are four state-of-the-art architectures that have been the best way to understand the significance of performance and light-weight architecture. 

\textbf{Deep CNN Architectures}: Deep CNNs (DCCN) have been introduced to generate local low-level and mid-level feature learning in the initial layers and high-level and global learning in the deeper layers. High-level features are the combination of low and mid-level features. VGG is one of those DCNNs that decreased the ImageNet error rate from 16\% to 7\% \cite{simonyan}. VGG expresses Visual Geometry Group. They employed small convolution filters of size 3$\times$3 in place of the large filters comprising 11$\times$11, 5$\times$5 filters and observed the significant performance with varying depth of the network. The use of small filters also reduced the no of parameters, thus decreasing the number of computations. ResNet50 is the next powerful and successful network after VGG, which stands for residual networks having 50 layers \cite{he}. ResNet was introduced to address two problems. The network has the ability to bypass a layer if the present layer could decrease performance and leads to overfitting, and this process is referred to as identity mapping. Another significance is to allow an alternate shortcut path for the gradient to flow that could avoid vanishing gradients problem \cite{dhillon}. ResNet decreased the error rate of 7\%(VGG) to 3.6\%. 

\textbf{Lightweight CNN Architectures}:  MobileNet V1 was an early attempt to introduce the lightweight model \cite{howard2}. The remarkable performance of this model was achieved by substituting the standard convolution operation with depthwise separable convolution. This enhances the feature representation, which makes the learning more efficient. The two primary components of depthwise separable convolution are depthwise and pointwise convolution, which are introduced to adjust channels and reduce parameters \cite{hua}. The stacking of these two components generates novelty in the model.  After this, another model proposed was Mobilenet V2 \cite{sandler}.  Mobilenet V2 is the extension of Mobilenet V1. Sandler and colleagues identified that non-linear mapping in lower-dimension increased information loss. To address this problem, a significant module was introduced with three consecutive operations. Initially, the dimension of feature maps is expanded using 1$\times$1 convolution, followed by, a depthwise convolution of 3$\times$3 to retain the abstract information. And in the last part, all the channels are condensed into a definite size using 1$\times$1 pointwise convolution. These transformations are processed into a bottleneck residual block which is the core processing unit in place of standard convolution. 

In our study, we use standard and depthwise convolution to enhance the performance and depthwise separable convolution to make the model time-efficient. The novelty of our work lies in the proposed pipeline to develop 2D plots from EEG signals having 3 RGB channels representing power spectral, which preserves the oscillatory information along with spatial position of electrodes. We classify three cognitive states comprising expert, non-expert meditative states, and control states (no prior experience of meditation). This paper discusses the following components a) Our proposed pipeline b) Experimentation on Dataset c) Comparative studies of ML and DL models d) Ablation Study

\section{Data and Methods}

\subsection{Data and Preprocessing}
We used two open-access repositories consisting of Himalayan yoga meditators and controls \cite{data1,braboszcz}. In this research, we used EEG data of 24 meditators and 12 control subjects. Twenty-four meditators were further divided into two groups comprising twelve experts and twelve non-experts. Data were captured using 64 channels Biosemi EEG system at the Meditation Research Institute (MRI) in Rishikesh, India. Experimental design and complete description are mentioned in the paper \cite{brandmeyer}. The expert group had an experience of a minimum of 2 hours of daily meditation for one year or longer, whereas non-experts were irregular in their practice. Control subjects had no prior meditation experience and were asked to pay attention to the breath’s sensations, including inhalation and exhalation. A recent study \cite{pandey} has shown the significant differences between expert and non-expert meditators and refers to as two distinct meditative states. Here, we refer to meditative and control states as cognitive states. As it includes cognitive components, such as involvement of attention during inhalation and exhalation of breathing, practitioners engaging in mantra meditation which enhances elements of sustained attention and language\cite{brandmeyer,lee,basso}. EEG data corresponding to breathing and meditation were extracted and preprocessed using Matlab, and EEGLAB software \cite{delorme}.  We classified three states emerging from expert, non-expert, and control groups, respectively.

EEG signals were downsampled at 256 Hz. A high pass linear FIR filter of 1 Hz was applied followed by removing the line noise artifacts at frequencies of 50,100,150,200, 250. Artifact correction and removal were performed using Artifact Subspace Reconstruction (ASR) method. Bad channels were removed and spherical interpolation was performed for reconstructing the removed signal, an essential step to retain the required signal. Data were re-referenced to average. Independent Components Analysis (ICA) was applied to classify the brain components and to remove the artifacts, including eye blink, muscle movement, signals generated from the heart, and other non-biological sources.

\subsection{Methods}
We divide the classification pipeline into two processing units. The first is to create power spectral density 2D plots from neural responses, known as “Neural Timeseries to 2D” and the second is to define the classification model term as “2D to Prediction” as shown in Fig. \ref{pipeline}.

\begin{enumerate}
    \item {\bf Neural Timeseries to 2D:} We divide the process of creating images from EEG signals into the following three steps.
    
\begin{enumerate}
\item \emph{Window Extraction}: EEG time-series signals are extracted into windows of 2, 4, and 6 seconds. For example, if a signal contains 24 seconds, we get 12, 6, and, 4 no of windows respectively. Varying window length identifies the information content and plays a significant role in discriminating the classes. In some applications, we have the luxury to extract varying windows sizes depending on the task.

    \item \emph{Power Spectral Analysis}: Power spectral density (PSD) is estimated for the extracted window using the Welch method \cite{pwelch}. Welch’s method is also known as the periodogram method for computing power spectrum, the time signal is divided into successive blocks followed by creating a periodogram for each block, and estimating the density by averaging. Oscillatory cortical activity related to meditation primarily observes in two frequency bands, theta (5-8Hz) and alpha (9-12Hz) \cite{brandmeyer}. These bands are further subdivided into theta1(5-6Hz),theta2(7-8 Hz), and alpha1(9-10Hz), alpha2(11-12Hz). For every channel, PSD is computed for the mentioned four bands. 
    
\item \emph{Topographic (2D- 3 Channel) Plot}: We use the topoplot function of the EEGLAB that transforms the 3D projection of electrodes in a 2-D circular view using interpolation on a fine cartesian grid \cite{topo}. Topographic plots were earlier implemented in Bashivan’s work \cite{bashivan}, and they combined plots from three bands to form one image of three channels, whereas we create one image of size 32 $\times$ 32 for each band having three RGB channels, and this might help to understand the significance of each band in the specific cognitive task. This plot preserves the relative distance between the neighboring electrodes and their underlying interaction, generating task-based latent representation using convolution. 
\end{enumerate}

\begin{figure*}
\centering
\begin{minipage}[b]{0.5\linewidth}
  \centering
  \centerline{\includegraphics[width=12cm,height=5cm]{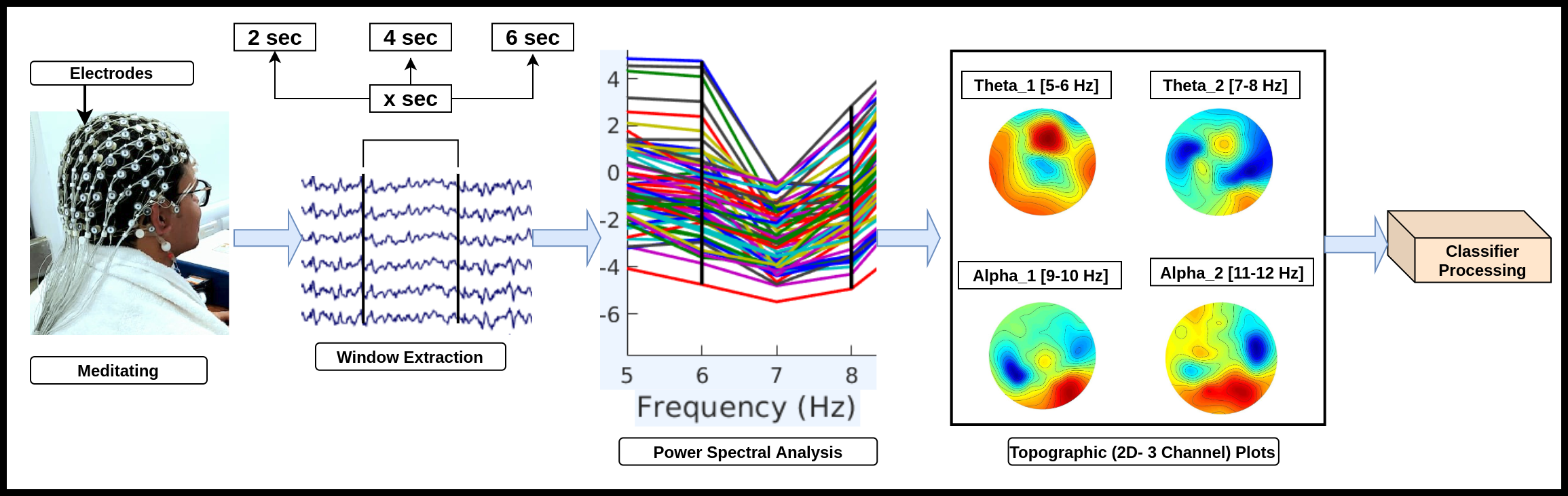}}
  \centerline{(a)Neural TimeSeries Signals to 2D: Images are formed using power spectral analysis.}\medskip
\end{minipage} \\
%
\begin{minipage}[b]{0.5\linewidth}
  \centering
  \centerline{\includegraphics[width=12cm, height =5cm   ]{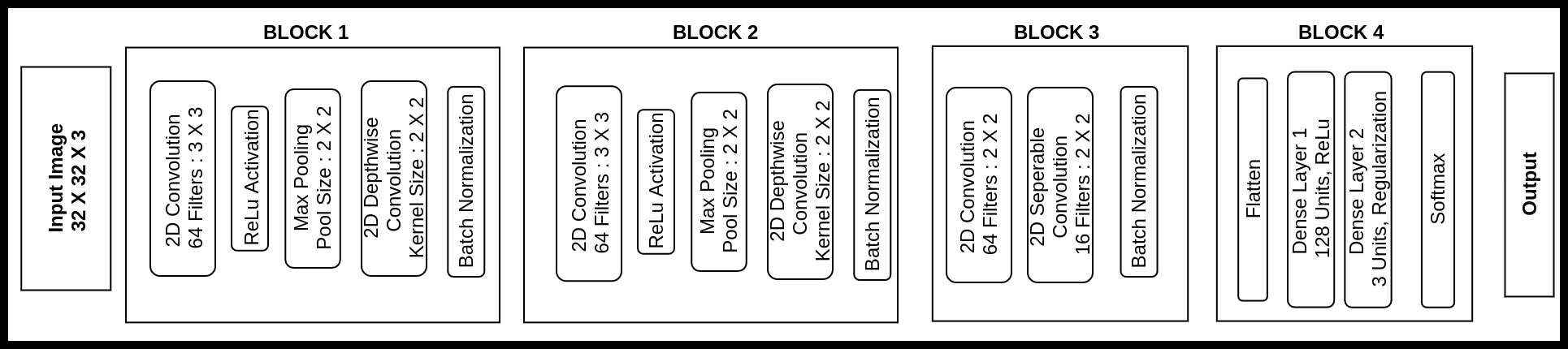}}
  \centerline{(b) 2D to Prediction: Images are fed into a deep learning network comprising of 4 blocks.}\medskip
\end{minipage}

  \caption{Proposed pipeline for classification}
  \label{pipeline}
\end{figure*}

\item {\bf 2D to Prediction}: Our model comprises 4 Blocks as shown in Fig. \ref{pipeline} (b). The first two blocks are introduced to capture deep feature information and the third block for reducing the computation.  The major components in these three blocks are regular Conv2D, depthwise spatial Conv2D, depthwise separable convolution, max pooling, ReLu(Rectified Linear Unit) activation function, and batch normalization.  

Conv2D to learn richer features at every spatial scale. Depthwise convolution which acts on each input channel separately and extremely efficient to generate succinct representation \cite{chollet}. Depthwise separable convolutions factorize a regular convolution into a depthwise convolution and a (1,1) convolution called a pointwise convolution \cite{howard}. This was initially introduced for generic object classification \cite{sifre} and later used in Inception models \cite{ioffe} to reduce the computation in the first few layers. Kernel sizes for the initial two blocks are 3$\times$3 and 2$\times$2 for standard 2D convolution with 64 filters and 2$\times$2 for Depthwise convolution. The third block comprises standard Conv2D and depthwise separable convolution of kernel size 2$\times$2 with 64 and 12 filters respectively. ReLu activation is introduced in the initial two layers to prepare the network to learn complex representations and to generate faster convergence and better efficiency \cite{nair}.  
The output of ReLu is processed by the Max Pooling operations for spatial sub-sampling, which downsample the feature maps by generating features in patches of the feature map \cite{chollet}.  Batch normalization is performed at last in the three blocks to minimize internal covariate shift, which subsequently accelerates the training of deep neural nets and enables higher learning rates \cite{ioffe}. 

Fourth block contains flatten, two dense layers and a softmax activation function. A dense layer is employed to combine all the features learned from previous layers where every input weight is directly connected to the output of the next layer in a feed-forward fashion. Since we are doing multiclass classification, the output layer has a softmax activation function. Softmax as an activation function is used because the model requirement is to predict only the specific class, which results in high probability. To find out the loss of the model, categorical cross-entropy is used to predict the probability of a class.

Deep learning models are trained using tensorflow(keras) \cite{keras} and machine learning algorithms employ scikit learn library \cite{pedregosa}. GPU NVIDIA GTX 1050(4GB RAM) are used for this study and the batch size are set to 30 because of memory constraints and kept the maximum epoch to 30 to maintain the timing constraint as well as to avoid overfitting and for optimal training and validation loss.
\end{enumerate}

\section{Results and Discussion}
This section discusses five measures a) performance of our model b) comparison of time and parameters c) training and validation loss d)visualization of layers e) ablation study

\subsection{Performance}
\textbf{Baseline Methods}: We compared the performance with commonly used classifiers and state of the art deep learning models. We trained six machine learning classifiers and tried several hyperparameters as shown in Table \ref{MLC}  and reported the maximum accuracy for all except NLSVM because it was around chance level. We experimented with four state of the art deep learning models VGG16\cite{simonyan}, ResNet50\cite{he}, MobileNet\cite{howard2}, MobileNetv2\cite{sandler} and keep block4 of our model as the last block in all the models. These models are best suited for classification tasks. We reported the cross-subject average accuracy using leave one out validation, eleven subjects from each condition were used for training and one subject from each condition was used for testing and then reported the average accuracy of 12 iterations. For example, in the 2-sec window,  9909 samples were used for training and 1101 for validation and 813 samples for testing, this was iterated for twelve times.

\begin{table}
\caption{Machine Learning Classifiers}
\centering
\resizebox{\columnwidth}{!}{%
\begin{tabular}{|l|l|} 
\hline
\textbf{Classifier}                                                         & \textbf{Parameters}                                                                                                            \\ 
\hline
\begin{tabular}[c]{@{}l@{}}Linear SVM \\(SVM)\end{tabular}                  & penalty parameter C values: 0.1,0.5,1.5,5,20,40,80,120,150~                                                                    \\ 
\hline
\begin{tabular}[c]{@{}l@{}}Non Linear\\SVM (NLSVM)~\end{tabular}            & \begin{tabular}[c]{@{}l@{}}kernel: 'rbf', upper bound on the fraction of margin errors \\nu : 0.3,0.4,0.5, 0.6\end{tabular}    \\ 
\hline
AdaBoost (AB)                                                               & \begin{tabular}[c]{@{}l@{}}algorithm = SAMME.R, no of estimators : 50, 100,200,250,\\300,400\end{tabular}                      \\ 
\hline
\begin{tabular}[c]{@{}l@{}}Logistic~\\Regression (LR)\end{tabular}          & regularisation : l1, solver: saga                                                                                              \\ 
\hline
\begin{tabular}[c]{@{}l@{}}Linear\\Discriminant\\Analysis(LDA)\end{tabular} & threshold rank estimation = 0.0001                                                                                              \\ 
\hline
\begin{tabular}[c]{@{}l@{}}Random~\\Forest( RF)\end{tabular}                & \begin{tabular}[c]{@{}l@{}}no of estimators: 50,100,150,200,250,300,400, min samples\\leaf:5, criterion: entropy\end{tabular}  \\
\hline
\end{tabular}
}
\label{MLC}
\end{table}

\begin{figure}
\centering
\includegraphics[width=8.5cm,height = 7cm]{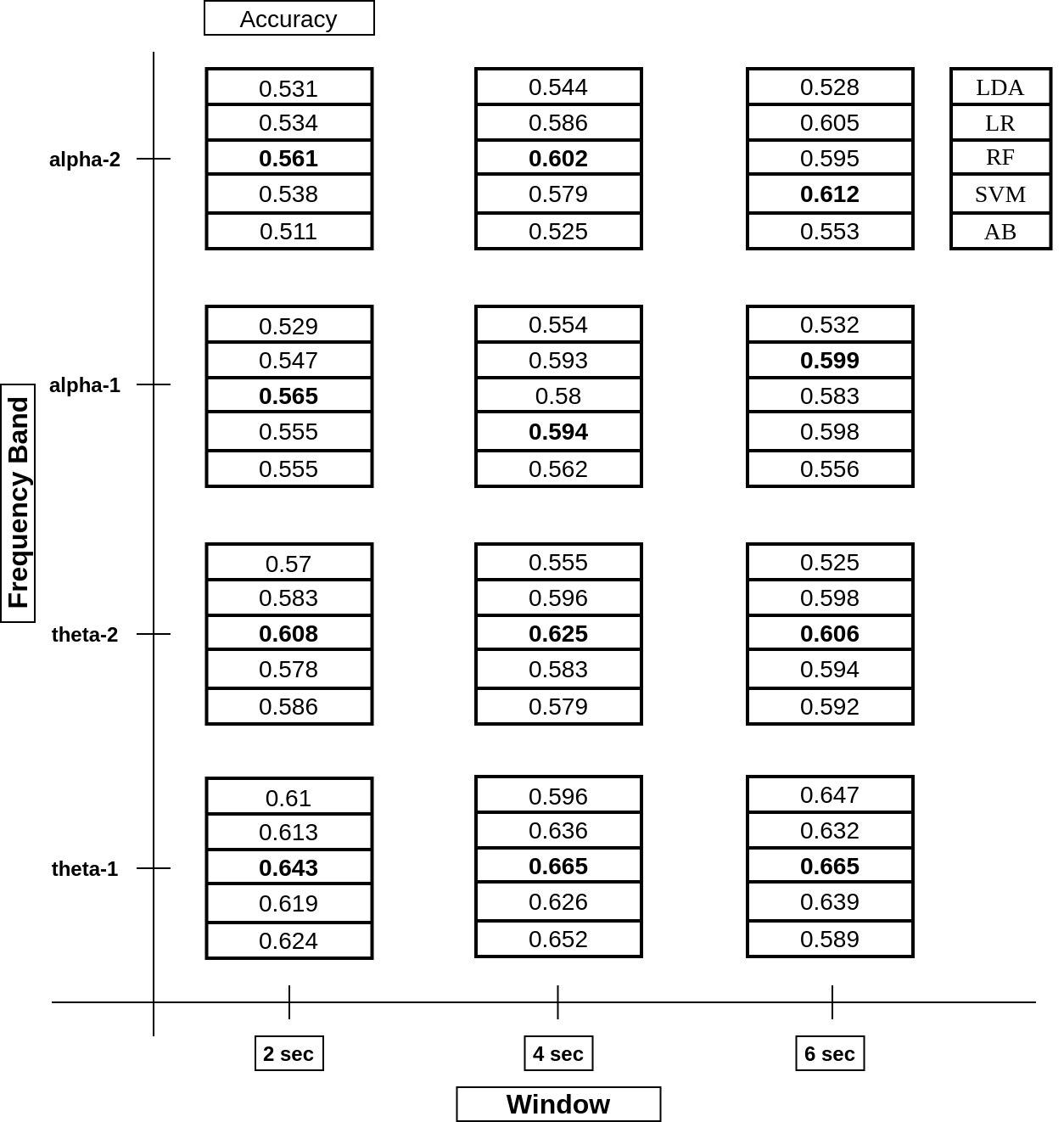}
\caption{Performance of ML Models: Classification of three cognitive states; expert, non-expert meditative states, and control. Each box indicates five accuracy values representing classifiers mentioned in the right top corner. Bold text represents the maximum value in the box.}
\label{ML}
\end{figure}

\begin{figure}
\centering
\includegraphics[width=9cm,height = 8cm]{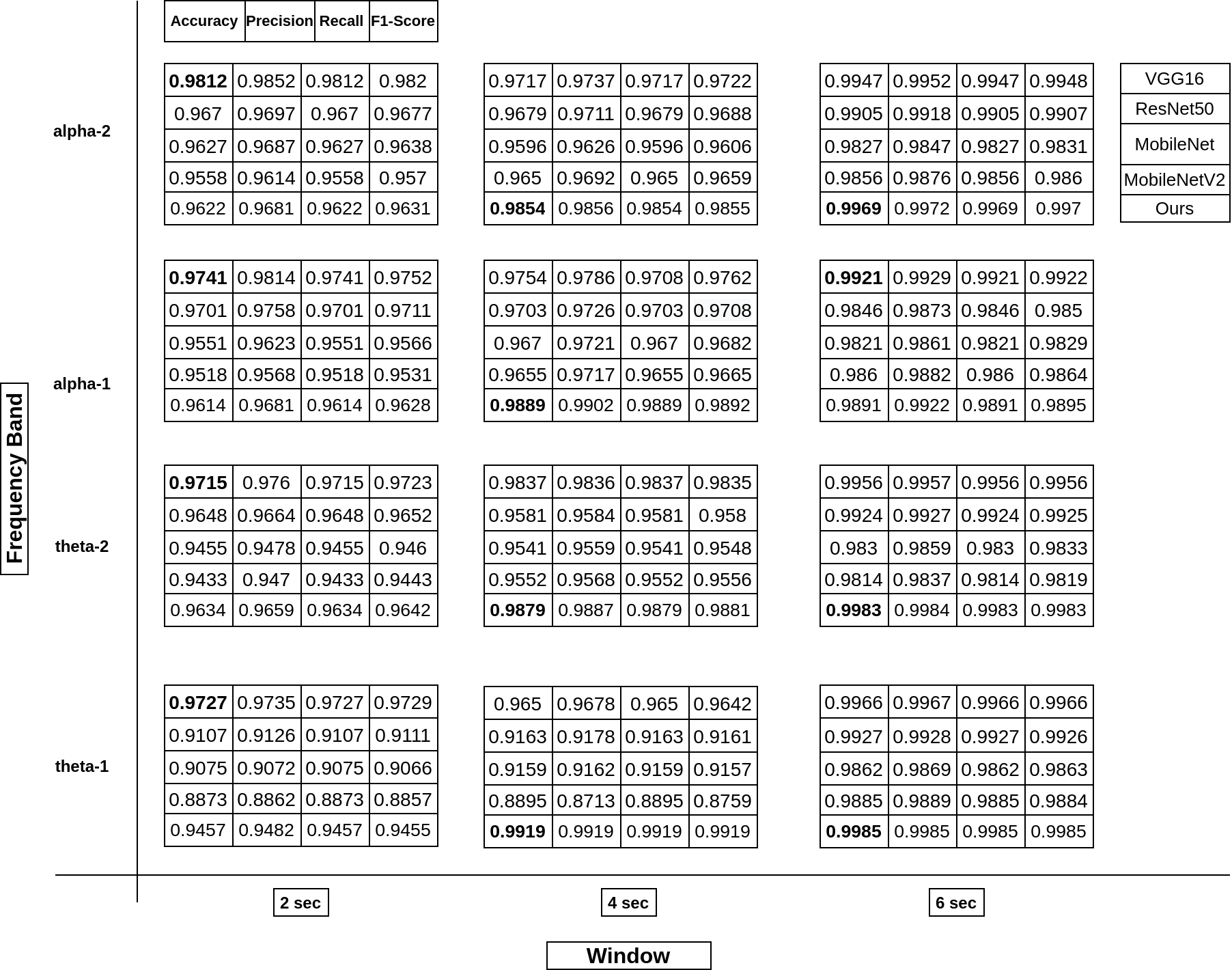}
\caption{Performance of CNNs: Each column of a box represents accuracy, precision, recall, and f1-score, and row indicates the CNN network mentioned on the right top corner.}
\label{DL}
\end{figure}

We obtained maximum accuracy 64.3\% in theta1, 60.8\% in theta2, 56.5\% in alpha1, and 56.1\% in alpha2 for a 2-sec window using Random Forest Classifier as shown in Fig. \ref{ML}. Deep learning models completely outperformed traditional ML techniques, we obtained 97.27\% in VGG16, 91.01\% in ResNet50, 90.75\% in MobileNet, 88.73\% in MobilenetV2 and 94.57\% in Brain2Depth(Ours) for theta1 in 2sec window as shown in Fig. \ref{DL}. The minimum accuracy in traditional classifiers might be due to utilizing the images directly for training by flattening the images into a features matrix. There might be an improvement if we could have tried with power spectral features but didn’t anticipate performance comparable with DL models. This cannot preserve the spatial position as the topo plot does.  However, when compared our model with DL models showed comparable performance in all windows.

\subsection{Parameters and Time}
Parameters define the complexity of a model, we tried to keep our model simple and explainable so that each block can be understood easily. We compared the parameters and time for training and testing. Brain2Depth demonstrated comparable performance while using parameters which were only 0.52\% of VGG16, 0.32\% of ResNet50, 2.28\% of MobileNet, and 3.16\% of MobileNetV2. The training and testing time of our model for 9909 samples and 813 samples were 84.882 and 0.152 seconds whereas the mobilenet took the minimum time in all four models, which were 214.989 and 0.44 seconds. Fast training may help to develop a model fast with millions of images and make a quick deployment. Minimum test time is most important in real-time prediction.

\begin{table}
\caption{The number of trainable parameters in each network with train and test time, respectively.}
\centering
\begin{tabular}{|l|l|l|l|l|} 
\hline
\textbf{Model} & \textbf{Parameters} & \textbf{Training Time(s)} & \textbf{Testing time(s)} & \begin{tabular}[c]{@{}l@{}}\textbf{Testing per}\\\textbf{sample(ms)}\end{tabular}  \\ 
\hline
VGG16          & 14780739            & 652                       & 0.473                    & 0.582                                                                              \\ 
\hline
ResNet50       & 23850371            & 847.483                   & 1.119                    & 1.376                                                                              \\ 
\hline
MobileNet      & 3360451             & 214.989                   & 0.44                     & 0.541                                                                              \\ 
\hline
MobileNetV2    & 2422339             & 258.503                   & 0.702                    & 0.863                                                                              \\ 
\hline
\textbf{Ours}  & \textbf{76627}      & \textbf{84.882}           & \textbf{0.152}           & \textbf{0.187}                                                                     \\
\hline
\end{tabular}
\label{Time}
\end{table}

\begin{table}

\caption{Test performance of each iteration on theta1 band of 2-sec window: Training performed on 33 subjects and testing on 3 subjects, including all conditions. Bold values represents two best performance.}
\centering
\resizebox{\columnwidth}{!}{%
\begin{tabular}{|l|l|l|l|l|l|l|l|l|l|l|l|l|} 
\hline
\textbf{Model} & \textbf{1}     & \textbf{2}     & \textbf{3}     & \textbf{4}     & \textbf{5}     & \textbf{6}     & \textbf{7}     & \textbf{8}     & \textbf{9}     & \textbf{10}    & \textbf{11}    & \textbf{12}     \\ 
\hline
VGG16          & \textbf{0.839} & \textbf{0.915} & \textbf{0.957} & \textbf{0.985} & \textbf{0.996} & \textbf{1}     & \textbf{0.998} & \textbf{0.988} & \textbf{0.998} & \textbf{0.999} & \textbf{0.997} & \textbf{1}      \\ 
\hline
RESNET50       & 0.701          & 0.617          & 0.834          & 0.934          & 0.975          & 0.989          & 0.975          & 0.939          & .989           & .998           & .988           & 0.99            \\ 
\hline
Mobilenet      & 0.619          & 0.689          & 0.786          & 0.935          & 0.974          & 0.996          & 0.981          & \textbf{0.946} & 0.986          & 0.993          & \textbf{0.998} & 0.988           \\ 
\hline
MobilenetV2    & 0.626          & 0.674          & 0.7            & 0.864          & 0.935          & 0.975          & 0.985          & 0.931          & 0.985          & 0.996          & 0.994          & 0.982           \\ 
\hline
Ours           & \textbf{0.797} & \textbf{0.852} & \textbf{0.865} & \textbf{0.946} & \textbf{0.987} & \textbf{0.997} & \textbf{0.998} & 0.915          & \textbf{1}     & \textbf{0.999} & 0.996          & \textbf{0.998}  \\
\hline
\end{tabular}
}
\label{sj}
\end{table}

\subsection{Training vs Performance Tradeoff}

\begin{figure}
\centering
\includegraphics[width=6.5cm,height = 5cm]{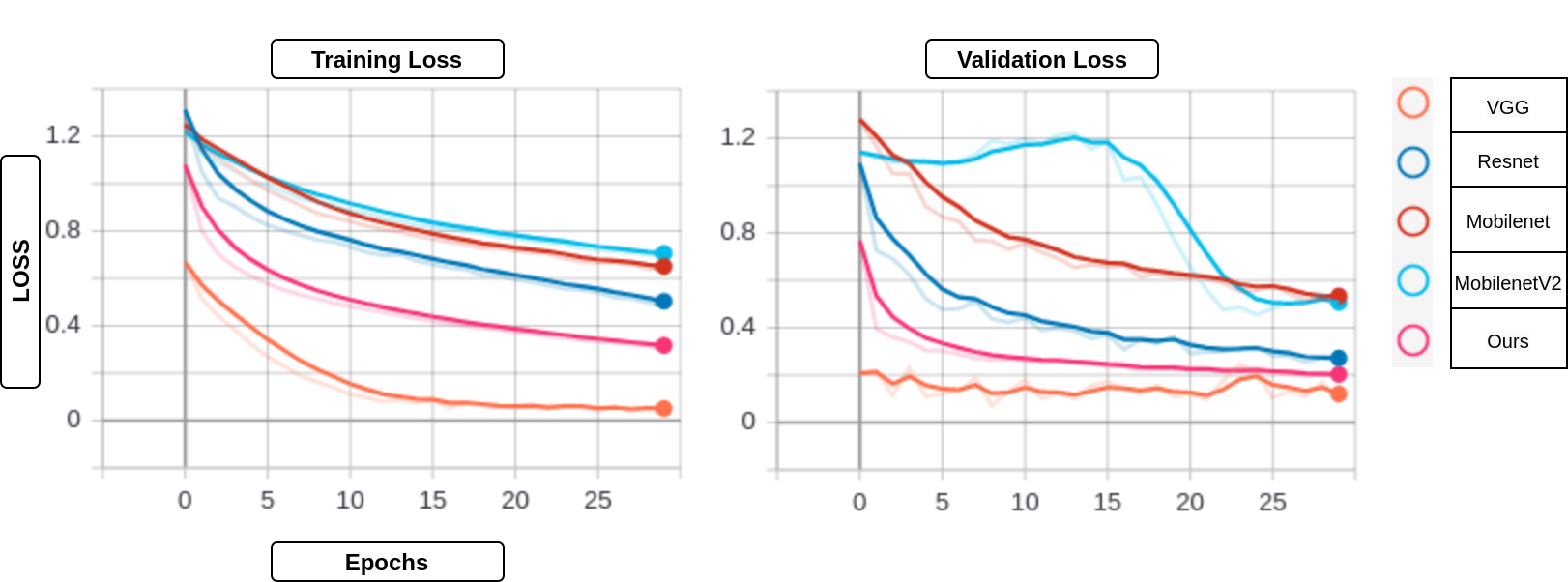}
\caption{Training and validation loss of the first iteration in theta1 band of 2-sec window with respect to all networks.} 
\label{TVL}
\end{figure}

ResNet, MobileNet, and MobileNetV2 performed a little low when compared with VGG and Brain2Depth for a 2-sec window in theta1 . We further explored the subjectwise performance to understand the differences. We found that accuracy dropped significantly for one and two iterations as shown in the Table \ref{sj}. In the next step, we investigated training and validation loss to verify whether this happened because of overfitting or underfitting. Fig. \ref{TVL} shows that training loss is high for three comparatively with others too. This shows that it may require more training rounds for this specific case however our model demonstrates that with moderate training it may perform well. And ResNet is a heavy architecture it may also require a large number of training samples. Hence with optimal time, our network can be trained efficiently even though the training sample can be noisy or varying samples.

\begin{figure}
\centering
\includegraphics[width=7.5cm, height=4cm]{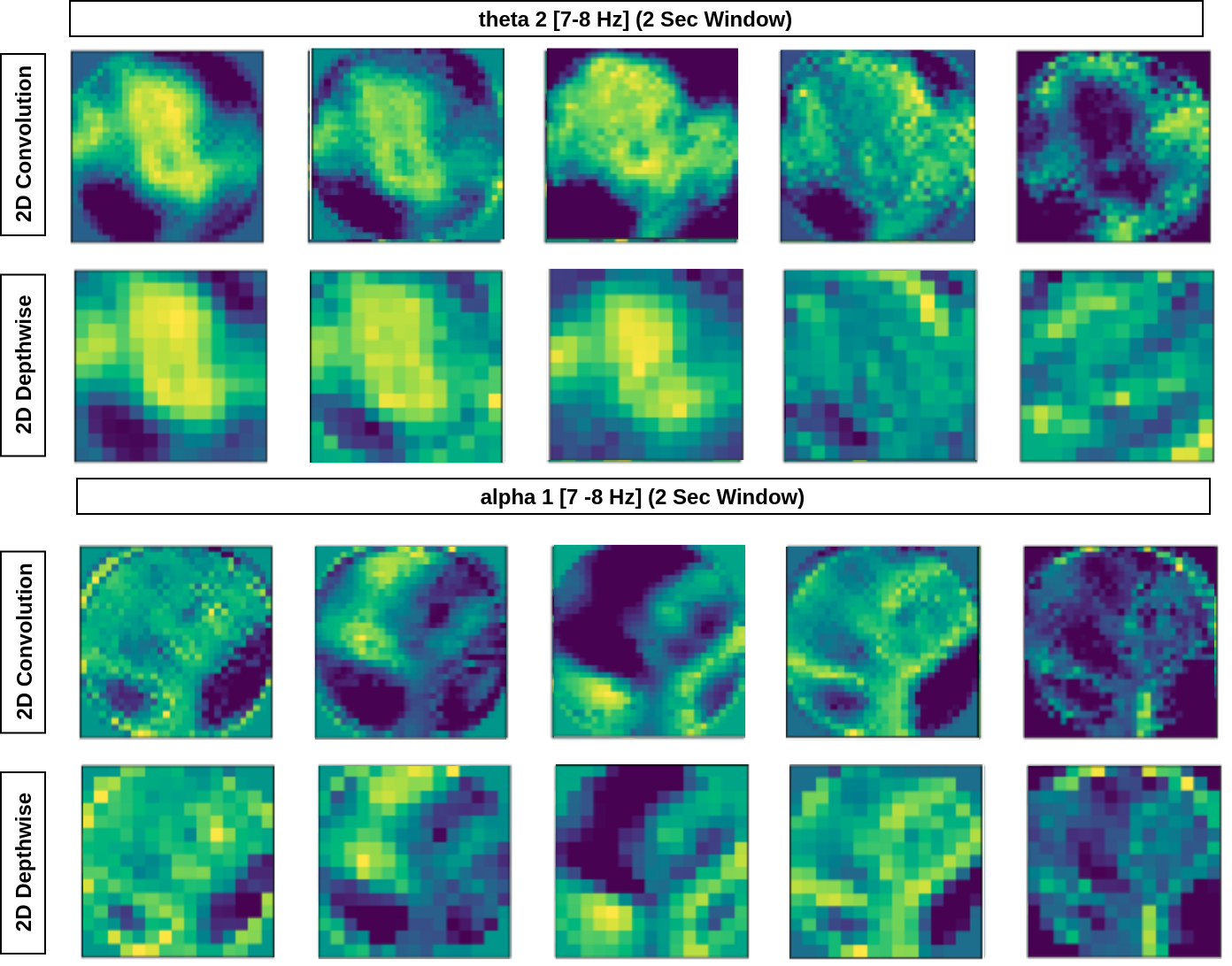}
\caption{Visualization of Block1 Layers: Intermediate representation of standard and depthwise convolution layers in theta2 (7-8 Hz) and alpha1 (9-10 Hz) bands for expert meditators.}
\label{V}
\end{figure}

\subsection{Visualization} 
We have visualized the intermediate CNN outputs in block1 for two frequency bands i.e. theta2 and alpha1. Fig. \ref{V} shows the outputs of five filters for 2D convolution and depthwise convolution in the expert condition. Layers have learned the different features, more specifically in theta2, the frontal region has shown the heightened contribution as compared with alpha1. This has also been studied in meditation research on the role of the frontal midline region in the experienced meditators \cite{brandmeyer}.

\begin{table}
\caption{Ablation Study: Performance evaluation on 2-sec window with (A) several changes in the number of filters and kernel sizes from block1 to block3. (B) change the order of Relu, Batch Normalization, and Max pooling (C) modify layer with another layer.}
\includegraphics[width=12.5cm]{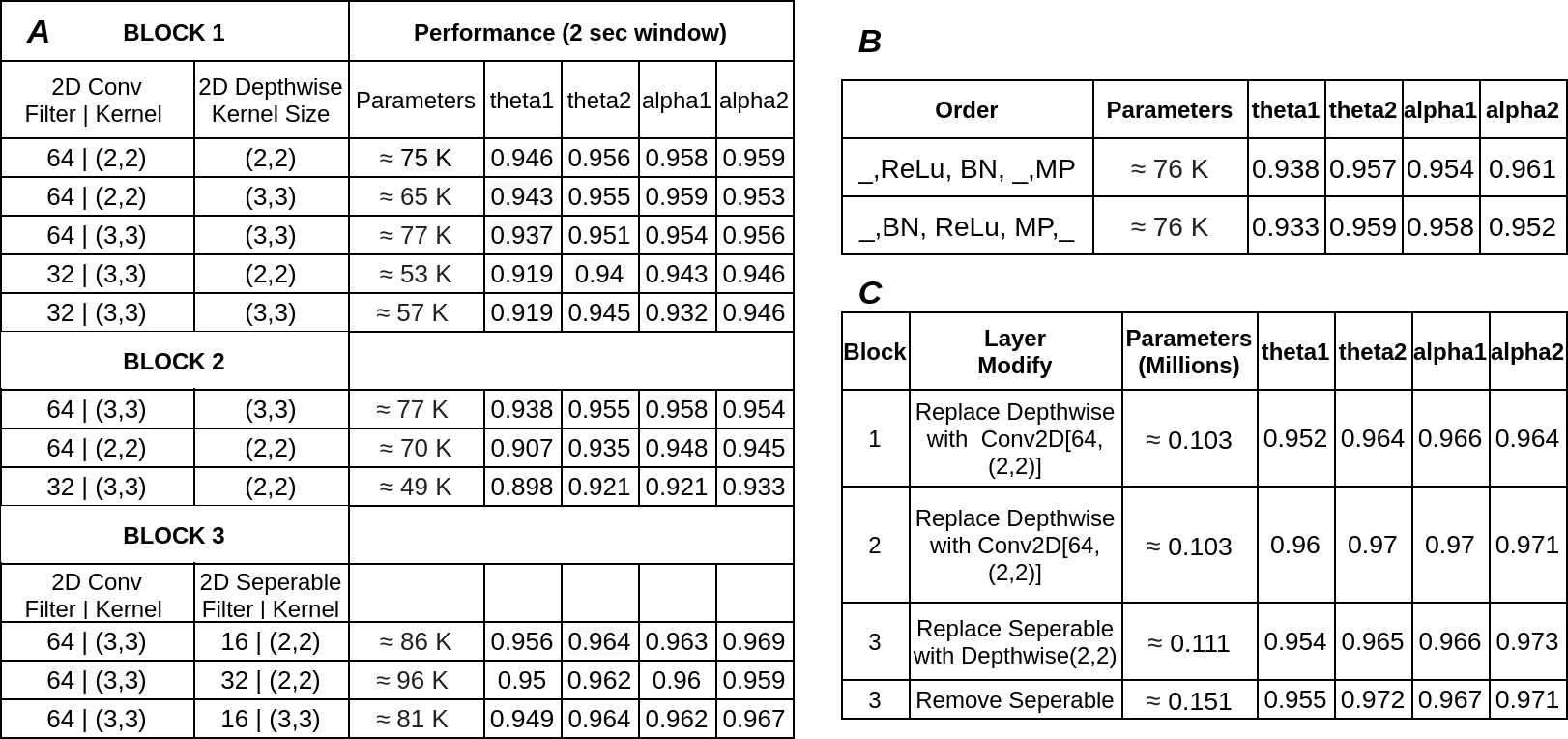}
\label{ABS}
\end{table}

\subsection{Ablation and Performance Studies}
We provide detailed ablation and performance studies. We exploited our model in the following three ways and reported the significant observations. 
\begin{enumerate}
    \item \emph{Variation in Filters:} Different level of granularity in the learning representation depends on the size and number of filters. Small-size filters generate fine-grained information, whereas large-size filters represent coarse-grained information. We varied the number of filters and their sizes into three blocks as shown in Table \ref{ABS} A. We kept the two blocks intact and changed the left block and observed the performance. No. of parameters ranged from 49K to 91K, and observed performance changes with varying parameters. Even with 53K parameters in block1, we found a significant performance above 90\% in all the bands.  This shows that our presented model can be further customized according to the need. 

 \item \emph{Swapping of components:} We swapped the positions of batch normalization, Relu, and max-pooling. We didn’t observe any significant differences as shown in Table \ref{ABS} B. 

\item \emph{Change in layers:}  In each block, we replaced one layer type with another layer as shown in Table \ref{ABS} C. We replaced depthwise convolution with standard convolution. We observed a slight improvement by (1-2)\% in all the frequency bands however parameters got double and increased the training timing by 7 seconds. Hence, this suggests that the model can be efficiently fine-tuned according to the availability of the resources and other constraints such as time. 
\end{enumerate}

\section{Conclusion}
This study exhibits a pipeline for the EEG classification task, incorporating steps to create topo images from EEG signals and a lightweight CNN model. In several medical domains, heavy deep architectures are not required, and a simple model is needed to produce similar results with less time. Our proposed study shows state-of-the-art performance while using only 0.52\% and 3.16\% parameters of the VGG and MobileNetV2 network, leading to a significant reduction in train and test time. Our model can be efficiently deployed in several real time protocols and effectively suited for resource constraint environment.

\section{Acknowledgement}
We thank SERB and PlayPower Labs for supporting PMRF Fellowship. We thank FICCI to facilitate this PMRF Fellowship. We thank Kushpal, Pragati Gupta and Nashra Ahmad for their valuable feedback. 

%
%
%
%

\bibliographystyle{splncs04}
\bibliography{refs}

\end{document}